\documentclass[12pt, a4paper]{article}
\usepackage[top=1in, bottom=1in, left=1in, right=1in]{geometry}

\usepackage{setspace}
\usepackage{paralist}
\usepackage{comment}
\usepackage[utf8]{inputenc}  
\usepackage[T1]{fontenc}     
\usepackage{ebgaramond}

\usepackage[ngerman, english]{babel}
\usepackage[iso, ngerman]{isodate}

\usepackage{mathtools}
\usepackage{amsfonts}       
\usepackage{amssymb}
\usepackage[cmintegrals,cmbraces]{newtxmath}
\usepackage{ebgaramond-maths}
\usepackage{nicefrac}
\usepackage{tabularx}
\usepackage{booktabs} 

\usepackage{hyperref}       
\usepackage{cleveref}
\usepackage{url}            
\usepackage{booktabs}       
\usepackage{microtype}      
\usepackage{xcolor}         
\usepackage{rotating}
\usepackage{graphicx}       
\usepackage{bbm}
\usepackage{tikz}
\usepackage{subcaption}
\usepackage{centernot}      
\usepackage{verbatim}       
\usepackage{csquotes}

\usepackage[numbers,square,comma]{natbib}
\title{Establishing Construct Validity in LLM Capability Benchmarks Requires Nomological Networks}

\author{Timo Freiesleben\thanks{Corresponding authors: \texttt{timo.freiesleben@lmu.de.}} 
}

\date{}

\begin{document}

\maketitle

\begin{center}
\begin{tabular}{c}
\footnotesize{Munich Center for Mathematical Philosophy (MCMP)}\\LMU Munich
\end{tabular}
\end{center}

\begin{abstract}
Recent work in machine learning increasingly attributes human-like capabilities such as reasoning or theory of mind to large language models (LLMs) on the basis of benchmark performance. This paper examines this practice through the lens of construct validity, understood as the problem of linking theoretical capabilities to their empirical measurements. It contrasts three influential frameworks: the nomological account developed by Cronbach and Meehl, the inferential account proposed by Messick and refined by Kane, and Borsboom’s causal account. I argue that the nomological account provides the most suitable foundation for current LLM capability research. It avoids the strong ontological commitments of the causal account while offering a more substantive framework for articulating construct meaning than the inferential account. I explore the conceptual implications of adopting the nomological account for LLM research through a concrete case: the assessment of reasoning capabilities in LLMs.
\end{abstract}

\textbf{Keywords:} {\em Construct Validity; Measurement; Benchmarking; Machine Learning}


\vspace{1.5cm}
\begin{quote}
\emph{``Measure what can be measured, and make measurable what cannot be measured.''}

\hfill --- Galileo Galilei
\end{quote}
\newpage
\section{Introduction}
\label{sec:intro}
Humans routinely attribute capabilities to one another. We say things like 'she is a sharp reasoner' or 'he has good moral intuitions'. Intuitively, we grasp what such claims mean: we recognize the patterns of behavior they point to and how these patterns tend to cohere with other capabilities a person may or may not possess. For example, by definition, we expect someone with strong reasoning skills to excel at solving mathematical problems or navigating complex legal cases. Beyond this, we also associate reasoning with related capabilities, such as broad knowledge or strong memory.

To measure complex capabilities like reasoning, psychologists and educational scientists have developed various tests, which typically consist of a range of concrete tasks like completing math sequences, logical syllogisms, or analogies. While such tests are not uncontroversial, evidence suggests that they provide a meaningful measure of what we mean by reasoning \citep{wilhelm2005measuring}.

Machine learning researchers have recently begun to transfer this capability-based language to the context of large language models (LLMs) \citep{harding2024machine}. They test LLMs for reasoning capability \citep{xu2025toward}, theory of mind \citep{strachan2024testing}, or even morality \citep{jiang2022machineslearnmoralitydelphi}. They do so in close analogy with human assessment, namely by designing concrete tests in the form of benchmarks, i.e. structured evaluation suites for machine learning models \citep{freiesleben2025benchmarking}. Attributing substantial capabilities such as reasoning to LLMs is risky: Humans naturally interpret terms like reasoning in the same way they interpret them when applied to other humans, drawing broad generalizations about behavior and inferring the presence of related capabilities \citep{watson2019rhetoric}. How can we be confident that our benchmark tests genuinely measure the relevant capability and thus warrant intuitive interpretation by human users?

In the context of human testing, a range of well-established standards for test design aims to ensure that tests measure the intended capability, known under the name \emph{construct validity}. The central task is to connect a latent construct, for example reasoning capability, with a concrete measurement, such as a reasoning test. Researchers investigating LLM capabilities have so far largely ignored these standards in designing their benchmarks and are only gradually becoming aware of their existence \citep{bean2025measuring,salaudeen2025measurement,alaa2025position}. But which philosophical framework of construct validity should LLM researchers adopt?

Within philosophical psychology, three highly influential accounts of construct validity can be distinguished \citep{colliver2012test}. They differ in their ontological commitments regarding the status of constructs, as well as in their views on how measurements should be validated and what kinds of inferences they support. The first is the \emph{nomological account} proposed by Cronbach and Meehl in 1955 \citep{cronbach1955construct}. On this view, a construct is defined by its role within a nomological web of relations to other constructs. Validity is established by testing whether measurements of the construct correlate with measurements of other constructs as specified by the nomological web; i.e. it takes a coherentist perspective. The second is the \emph{inferential account} advanced by Messick in 1989 \citep{messick1989meaning,messick1995standards} and later refined by Kane \citep{Kane2013}, which, rather than asking whether a measurement truly captures a construct, understands construct validity in terms of the evidence and assumptions that link a measurement to its interpretation and use. The most recent is the \emph{causal account} defended by Borsboom in 2004 \citep{borsboom2004concept,borsboom2005measuring}. This view requires the ontological existence of the construct of interest and demands that it plays a causal role in producing the observed behavior.

In this paper, I argue that the nomological account developed by Cronbach and Meehl provides the most suitable theoretical framework for current LLM capability research (\Cref{sec:nomologicalBest}). Compared to Borsboom’s causal account, it avoids strong ontological commitments about the real existence of constructs, commitments that are difficult to defend in the case of LLMs. Unlike the inferential account, it offers concrete guidance on how to theorize complex constructs. I then analyze the conceptual implications of adopting the nomological account in LLM research by applying it to the assessment of reasoning capabilities (\Cref{sec:theory}).

\section{The nomological account is the best available framework for conceptualizing LLM capabilities}
\label{sec:nomologicalBest}
There are three major accounts of construct validity in psychological and educational testing \citep{colliver2012test} that are potential candidates for evaluating LLM capabilities: the nomological account, the inferential account, and the causal account. In the following, I analyze the suitability of each as a framework for construct validity in LLM capability testing in reverse historical order.

\subsection{The causal account is ontologically demanding for LLMs}
\label{subsec:causal}
Borsboom's causal account of construct validity is conceptually straightforward. For a measurement to validly measure a construct, two conditions must be satisfied: the construct must \emph{exist}, and variations in measurement outcomes must be \emph{caused} by variations in the state of that construct \citep{borsboom2004concept,borsboom2005measuring}. On this view, construct validity is established by demonstrating a genuine causal mechanism linking the construct to test scores. Validity thus becomes a scientific claim about underlying causal structure, grounded in a realist ontology of psychological attributes.

This ontological perspective may be defensible in the context of human cognition, where constructs such as reasoning or theory of mind are at least plausibly connected to biological and cognitive mechanisms \citep{bechtel2007mental}.\footnote{Even in the case of humans, the ontological status of constructs remains controversial \citep{asay2018role,oude2020folk}.} For LLMs, however, this realist stance is difficult to defend. It is unclear what it would mean for a model to \emph{possess} reasoning, theory of mind, or morality as internal attributes, rather than merely to \emph{exhibit behavioral patterns} that resemble those associated with these constructs in humans. In any case, inferring the existence of a construct from observed behavioral patterns requires particularly strong evidence \citep{Kane2013}. The burden would fall on researchers defending the causal account to provide this evidence.

One possible response from defenders of the causal account is to appeal to mechanistic interpretability research \citep{sharkey2025open} and argue that capabilities are realized in identifiable internal components of LLMs, thereby supporting their existence in Borsboom’s realist sense. On this view, if specific circuits or subnetworks can be shown to causally produce particular behavioral regularities, then constructs correspond to genuine causal structures within the model. However, the success of mechanistic interpretability in isolating psychologically meaningful constructs, especially higher-level constructs, often remains fragmentary \citep{kastner2024explaining,freiesleben2025artificial,milliere2024philosophical}. Existing results mainly identify mechanisms for relatively narrow behaviors, e.g. related to concrete locations like Paris \citep{meng2022locating}, objects like the Golden Gate Bridge \citep{templeton2024scaling}, or mathematical operations like addition \citep{kantamneni2025language}. While such examples may provide partial support for a realist interpretation, it remains unconvincing unless mechanistic interpretability demonstrates robust mappings between internal structures and behavioral patterns associated with complex capabilities such as reasoning.

Alternatively, proponents of the causal account may conclude that if no construct exists in their strong realist sense, then no valid measurement of it is possible. On this view, if the causal account cannot be applied, we should refrain from speaking of measuring capabilities through benchmarks altogether; capability ascriptions would then be reserved for humans. This position, however, seems overly restrictive. The practical relevance of capability language in the context of LLMs is not tied to metaphysical realism about constructs, but to the usefulness of such language for humans in describing, predicting, and evaluating LLM behavior.

Finally, defenders of the causal account may adopt a weaker form of causal realism. Instead of assuming the existence of capabilities within the LLM, they may identify LLM capabilities with their corresponding behavior. However, different behaviors generally do not \emph{cause} one another. An LLM’s behavior in theory of mind tasks may be associated with its behavior in reasoning tasks, but it is certainly not caused by it. Stripped of its causal interpretation, theorizing constructs through structural causal models \citep{borsboom2005measuring} reduces to Bayesian networks \citep{bovens2004bayesian}. In this weaker form, causal realism ultimately theorizes constructs in much the same way as the nomological account \citep{briganti2023tutorial}.

\subsection{The inferential account provides limited resources to theorize LLM capabilities}
\label{subsec:argument}
Messick's and Kane's inferential conception of construct validity differs fundamentally from Borsboom's causal approach. Rather than asking the ontological question whether a measurement is caused by a latent but real construct, they focus on the epistemological question whether inferences drawn from a measurement are warranted. Because the validity of inferences depends on many factors, including background assumptions, related evidence, and the ethical consequences of score use, Messick argues that construct validity ``undergirds all score-based interpretations'' \citep{messick1989meaning}. The central goal becomes justifying interpretive claims about what test scores mean and how they may be used.

This shift has a clear advantage over the causal account: it avoids metaphysical commitments regarding the existence of constructs. Instead of asking whether a construct is real, Messick and Kane ask whether score interpretations are adequately supported by evidence. What in the causal framework requires ontological justification can here be defended pragmatically. However, as argued by Borsboom \citep{borsboom2004concept} and Lissitz et al. \citep{lissitz2007suggested}, this expansion renders construct validity extremely broad. The measurement process becomes intertwined with the ethical and practical consequences of test use, making it difficult to determine where construct validity is actually established, as it becomes dispersed across a wide range of inferential considerations. More importantly, the inferential account offers limited resources for even theorizing the construct. Because it focuses primarily on downstream inferences, the question of what exactly is being measured remains comparatively undertheorized. This limitation is particularly problematic for complex and abstract constructs such as reasoning, theory of mind, or morality, which LLM researchers aim to assess. 

For complex constructs, meaning cannot be established solely through inferential practices---a point even acknowledged by inferentialists such as Kane \citep{Kane2013}. Kane distinguishes between observable attributes and theoretical constructs, noting that inferences about theoretical constructs typically require nomological networks (introduced in \Cref{subsec:nomological}) as their theoretical foundation. Accordingly, it has been proposed that inferentialist, causal, and nomological accounts should be seen as complementary rather than fundamentally distinct approaches to construct validity, with inferentialist accounts focusing on the pragmatic dimension, nomological networks providing the theoretical grounding, and causal accounts offering a causal interpretation where possible \citep{hood2009validity,Kane2013}. Recent frameworks for benchmark validation in machine learning have largely adopted an inferential perspective, but they either focus on narrower constructs, such as weather forecasting \citep{freiesleben2025benchmarking}, or integrate nomological networks into their inferential framework to theorize more complex constructs \citep{salaudeen2025measurement}.

\subsection{Nomological Networks as a middle ground between causal realism and pure pragmatism}
\label{subsec:nomological}
Cronbach and Meehl originally introduced construct validity through the notion of a nomological network: a structured web of theoretical constructs, observable variables, and empirically testable relations. On this view, a construct is defined neither by its real-world referent, as in the causal account, nor by isolated inferential claims, as in the inferential account, but by its position within a coherent system of laws, correlations, and theoretical expectations.

The motivation for this proposal lies in the tradition of logical empiricism, which moved away from rigid \emph{point-by-point} definitions toward a more holistic conception of theory. Quine’s famous \emph{Web of Belief} illustrates this idea: the meaning of any single term is determined by its inferential relations to other terms and its distance from the sensory periphery \citep{quine1978web,quine1980logical}. To revise the definition of \emph{reasoning} in an LLM setting is therefore to adjust its relations to neighboring constructs such as memory or pattern recognition. Similarly, Hempel described scientific theories as complex networks of interconnected elements \citep{hempel1952fundamentals}. Theoretical constructs function as \emph{knots}, while logical and law-like relations serve as the \emph{threads} connecting them. Although the network is anchored in observable data, the constructs themselves derive their identity from their functional role within the system.

Because meaning is defined relationally within a nomological network, the ontological critique against the causal account does not apply. Nevertheless, Borsboom contends that avoiding metaphysical commitments comes at a cost: instead of providing \emph{reference} to a real but unobservable entity, the nomological account merely provides \emph{meaning} by situating terms within a semantic web. He argues that this project is bound to fail because ``Few, if any, theoretical terms in psychology can be unambiguously identified in this way.'' \citep{borsboom2004concept} However, LLM research suggests that in many practical contexts terms can be characterized relationally \citep{poschmann2024vector}. One might respond that Borsboom's critique targets the relatively simple nomological networks employed in psychological practice. But if so, the limitation lies not in the nomological approach itself, but in its empirical implementation. While constructing rich nomological networks in psychology is constrained by the difficulty of testing the same participants across many constructs, benchmarking additional tasks on the same LLM scales easily.

Above, I argued that the inferential account presupposes a sufficiently articulated construct to anchor its interpretive argument; it offers comparatively little guidance for theorizing the construct in the first place. By contrast, Cronbach and Meehl's nomological approach fills this gap by systematically embedding a construct within a theoretical network of relations. Rather than focusing on the \emph{inferences} warranted by benchmark scores, the nomological account places the \emph{meaning} of the construct itself at the center of attention. For emerging domains such as LLM capability research, where the meaning of constructs such as reasoning remains contested, this difference in emphasis is crucial. In such contexts, prioritizing the clarification of construct meaning is the more promising strategy. However, if LLM capabilities become more extensively theorized in the future and attention shifts to the legal or ethical implications of benchmark scores, the nomological account would benefit from incorporating insights from the inferential tradition \citep{kane2001current,hood2009validity,Kane2013}.

The main challenge lies in constructing a nomological network, especially for complex constructs such as reasoning. Here, the best available scientific theories should guide construction. For psychological constructs, one promising approach is to integrate established nomological models from psychology with additional relational information, for example, by extracting structural relations from LLM semantic vector spaces \citep{lukyanenko2024integrating} or learning them directly from data \citep{epskamp2018estimating}. Relying on nomological networks established in \emph{human} psychology in testing \emph{LLMs} might be seen as problematic. However, it helps ensure that the meaning of the constructs remains aligned with their use in interpersonal contexts (discussed in more depth in \Cref{subsec:prior}).

In sum, Cronbach and Meehl's nomological network account offers a compelling middle ground for LLM capability testing. It avoids Borsboom’s strong realist commitments while preserving a substantive notion of construct meaning that goes beyond Messick and Kane's inferentialism. To explore the conceptual implications of the nomological account for LLM research in greater depth, I now turn to a concrete case: the assessment of reasoning capabilities.

\section{Implications of the nomological account for LLM benchmark design}
\label{sec:theory}
In current practice, LLM researchers often begin by designing benchmarks for complex capabilities such as reasoning \citep{xu2025toward}, theory of mind \citep{strachan2024testing}, or even morality \citep{jiang2022machineslearnmoralitydelphi}. In psychology and educational science, however, directly designing tests for complex constructs is considered ``not good practice'' \citep{wilhelm2005measuring}. The danger lies in \emph{construct underrepresentation}, i.e., relevant parts of the construct may not be captured by the test \citep{strauss2009construct}, a central threat also in current LLM research, where data is often selected based on availability rather than representativeness \citep{Suehr2025}. To prevent construct underrepresentation, the nomological account recommends first designing a nomological network before constructing any test.

\subsection{Researchers should design nomological networks prior to benchmarks}
\label{subsec:prior}
\begin{figure}[p]
    \centering
    \includegraphics[height=0.8\textheight,keepaspectratio]{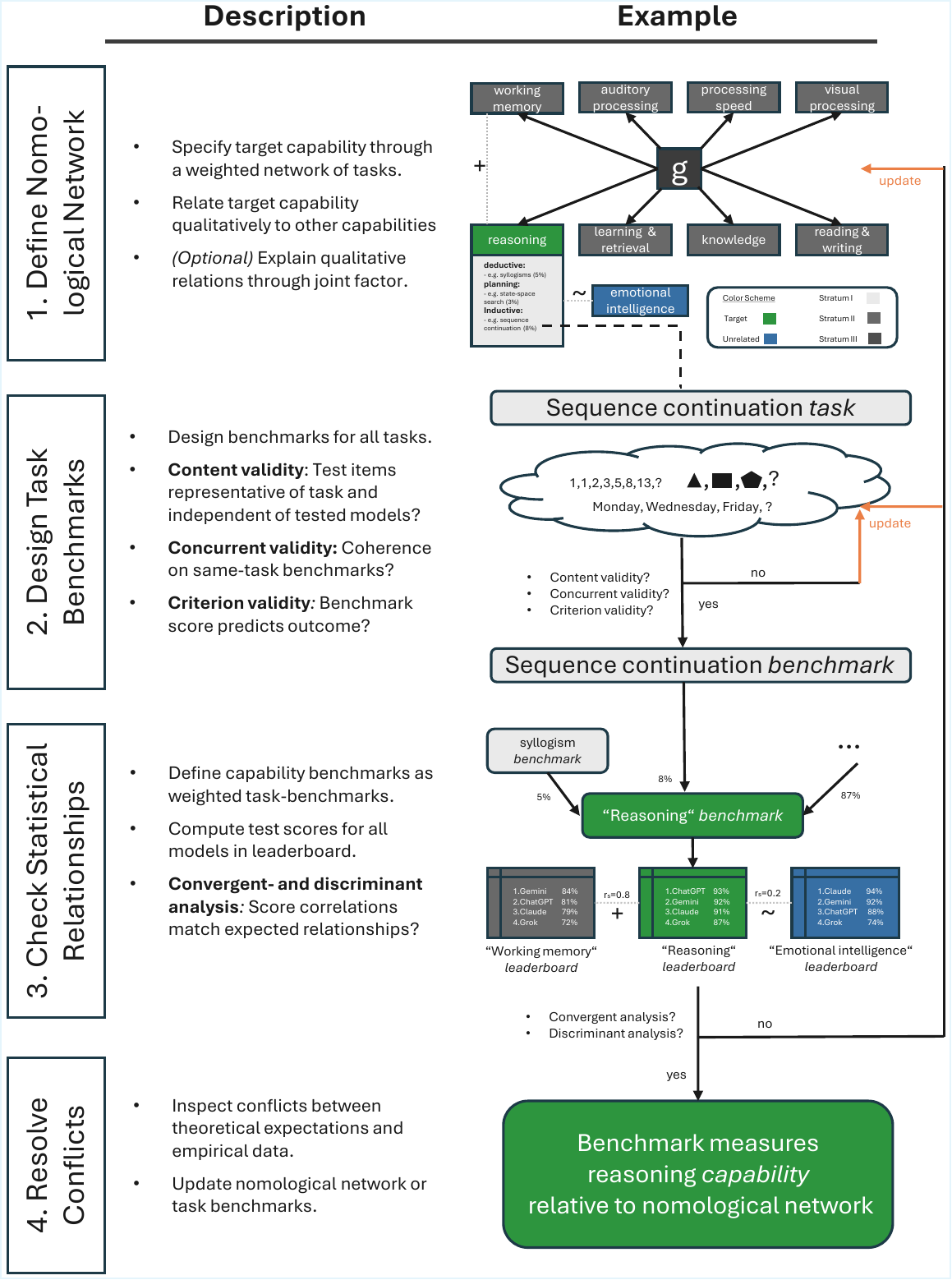}
    \caption{The nomological account structures the benchmark design process into four steps. The right-hand side illustrates the framework using reasoning as an example, with the Cattell–Horn–Carroll theory \citep{carroll1993human,mcgrew2005cattell} as the nomological network. To keep the example readable, only a subset of reasoning tasks is described and only two qualitative relationships are specified: ‘+’ indicates an expected positive relationship between reasoning and working memory \citep{kyllonen1990reasoning}, and ‘\textasciitilde’ denotes an expected independence between emotional intelligence and reasoning \citep{mayer2008emotional}.}
    \label{fig:overview}
\end{figure}

Consider the example of \emph{reasoning}. Designing a nomological network for reasoning requires specifying the qualitative relationships between reasoning and simple tasks such as continuing mathematical sequences, as well as other complex capabilities like working memory. In psychological testing, many constructs already have nomological networks supported both theoretically and empirically; reasoning is such an example \citep{wilhelm2005measuring}. Reasoning is commonly embedded in the Cattell–Horn–Carroll (CHC) theory of cognitive abilities \citep{carroll1993human}, widely regarded as the most empirically supported framework in contemporary psychometrics \citep{schneider2018cattell,mcgrew2005cattell}. CHC theory specifies a hierarchical structure of cognitive abilities (see \Cref{fig:overview}) organized into three strata \citep{carroll1993human}. Stratum I contains narrow \emph{tasks}, Stratum II broad \emph{capabilities}, and Stratum III a general \emph{cognitive ability factor}. Within this framework, reasoning is a capability.

In CHC, the meaning of \emph{reasoning} is partly determined by its relations to other capabilities, such as working memory capacity, processing speed, visual processing, and long-term retrieval. This interrelatedness is not accidental but theoretically structured. The correlation among capabilities is explained in CHC theory by reference to a higher-order construct: the $g$-factor, often interpreted as general intelligence. Although the ontological status of $g$ remains debated, most contemporary psychologists accept that modeling a higher-order general factor provides a coherent representation of observed covariance patterns \citep{rindermann2020survey}. From a nomological perspective, the key point is not metaphysical commitment to a reified $g$, but the structural constraint it imposes: capabilities are expected to correlate in systematic ways. Tasks, by contrast, are narrower; they are constituents of capabilities. For reasoning, these include inductive reasoning tasks such as sequence continuation, deductive reasoning tasks such as logical syllogisms, or planning tasks such as multistep search problems. While further decomposition is possible, tasks are typically treated as sufficiently specific to be operationalized through concrete tests \citep{cronbach1955construct}. In CHC, capabilities are modeled as weighted composites of tasks.

This hierarchical architecture already reveals something philosophically significant that goes beyond current benchmarking practice: reasoning is not identified with any single task but with a structured pattern of relations to narrow tasks and neighboring capabilities. A benchmark that ignores this structure risks mischaracterizing what it claims to measure. 

But should LLM researchers adopt CHC theory when testing LLM capabilities? CHC is, after all, a theory of \emph{human} cognition, not of \emph{LLM} cognition. Applying it to LLMs presupposes that its structural relations are transferable across radically different substrates—a non-trivial assumption. I believe that beginning with a theoretically mature and empirically supported human framework provides a principled starting point and ensures that our technical definitions align with the human tendency to interpret terms such as \emph{reasoning} or \emph{morality} as we do in interpersonal contexts. Nevertheless, a nomological network for LLM reasoning may ultimately diverge from the human CHC model, simply because reasoning in LLMs may occupy a different place within the broader architecture of capabilities than it does in human cognition.

\subsection{Task benchmarks require empirical and theoretical validation}

Assume the LLM researcher settles on a nomological network, here CHC (\Cref{fig:overview}). The second step in the nomological account is to design tests for \emph{tasks} (Stratum I). Consider, for example, the task of \emph{sequence continuation}. It can be described as predicting a symbol based on a sequence of other symbols, for instance, filling the question mark in the sequence 1, 1, 2, 3, 5, 8, 13, ?. Designing a test for the task means specifying a range of sequence continuation problems, the so-called \emph{test set}, consisting of concrete \emph{test items}. Tests also require a notion of error by assigning numerical values to predictions. For example, the answer 21, correctly continuing the Fibonacci sequence, could be assigned an error of 0, whereas any other answer could be assigned an error of 1. Furthermore, tests require test-takers and an overall test score obtained by averaging the errors across all test items.

These components of psychological tests match closely the components of machine learning benchmarks \citep{freiesleben2025benchmarking,Hardt2025}. However, although psychological tests and benchmarks are structurally similar, the validity theory surrounding psychological testing is substantially more mature. This becomes particularly evident when interpreting test scores: LLM researchers often interpret benchmark scores naively as task performance, whereas psychologists have developed a rich repertoire of conditions required for this inferential step.

The only validity conditions well known in LLM research are statistical in nature \citep{Hardt2025}: 1) test items should be sampled independently and identically (i.i.d.) from an underlying distribution, or at least form a representative sample of a population \citep{holtgen2024we}; 2) the test set should be independent from the test-takers; 3) the test set should be sufficiently large. These conditions are also recognized in the nomological account, where they constitute one aspect of \emph{content validity}\footnote{In other accounts they appear under different names—for example, as \emph{generalization} in Kane’s framework \citep{Kane2013}, \emph{statistical assumptions} in Borsboom’s framework \citep{borsboom2005measuring}, or \emph{internal validity conditions} in Freiesleben and Zezulka \citep{freiesleben2025benchmarking}.} and safeguard against selecting biased samples.

However, content validity in Cronbach and Meehl’s account encompasses more than statistical criteria. It also requires theoretical justification for the chosen task formalizations, test items, and error definitions, considerations that should likewise be addressed in LLM benchmarks. Beyond content validity, Cronbach and Meehl recommend evaluating the \emph{concurrent validity} of tests. Translated to benchmarks, this means that if two independently constructed benchmarks both claim to measure sequence continuation with high content validity, their scores should correlate strongly. Recently, increasing attention has been paid to this issue within machine learning under the label \emph{external validity} \citep{recht2019imagenet,Hardt2025,freiesleben2025benchmarking}. Similarly, when benchmark scores serve as proxies of real-world abilities, for example, selecting an LLM to fill tax forms based on reasoning scores, the nomological account requires verifying that the relevant skill correlates strongly with the score, i.e. \emph{criterion validity}.

One might conclude that established psychological tests such as the Woodcock–Johnson IV, the Wechsler scales, or the Stanford–Binet, since they satisfy these conditions, could simply be adopted for LLM testing. Indeed, they can guide test design, but many validity conditions depend partly on the test-taker. Consider statistical condition 2) above: LLMs are trained on vast corpora of data that may contain many psychological test items. In such cases, the test may measure memorization of correct answers rather than skill on the task. This phenomenon is known as \emph{data contamination} and has recently received increasing attention in the literature \citep{milliere2024philosophical,Suehr2025}.

\subsection{Capability benchmarks should undergo convergent and discriminant analysis}

The next step in the nomological account is to define capability benchmark scores for all capabilities in the nomological network. The benchmark score for reasoning capability is defined as the weighted average of the task benchmark scores connected to reasoning in CHC \citep{carroll1993human}.

Aggregation alone does not secure construct validity. From a nomological perspective, reasoning is not merely a sum of tasks but occupies a specific position within a broader network of capabilities like working memory or processing speed. If an LLM scores extremely high on reasoning but very low on working memory, as some studies suggest \citep{zhang2024working,hendrycks2025definition}, this creates a theoretical–empirical mismatch (see \Cref{subsec:resolving}).

This implies that LLM researchers must examine whether the theoretical relations posited in the nomological network align with the empirical relations observed in benchmark data \citep{carroll1993human}. In Cronbach and Meehl, this process is called \emph{convergent} and \emph{discriminant analysis}. Convergent analysis checks whether reasoning scores correlate positively/negatively with theoretically related capabilities such as working memory. Discriminant analysis checks that reasoning scores are uncorrelated with capabilities that are theoretically distinct, such as social intelligence. Discriminant analysis helps researchers to prevent \emph{construct overrepresentation} \citep{strauss2009construct}, i.e. benchmarks should not measure capabilities unrelated to the target capability, empirical research indicates that many current benchmarks suffer from this threat \citep{bean2025measuring}.

\subsection{Resolving conflicts between theory and measurement is difficult}
\label{subsec:resolving}
Empirical results and the theoretical expectations encoded in the nomological network may come into conflict. There is no straightforward way to resolve such tensions. We are confronted with a version of the Duhem–Quine problem \citep{van1976two,duhem1954aim,balashov1994duhem}: as Simms puts it, it remains unclear whether ``(i) the measure does not adequately measure the target construct, (ii) the theory requires modification, or (iii) some combination of both.'' \citep{simms2008classical} Applied to reasoning benchmarks, we may need either to question the adequacy of our task measurements or to reconsider aspects of the nomological network itself.

In the context of LLM research, I believe that initial doubt should often fall on the benchmarks. Benchmark design in machine learning is still underexplored, only gradually developing rigorous validation practices \citep{Hardt2025}. Theoretical–empirical mismatches will frequently indicate weaknesses in benchmark design. Established psychometric standards, such as criteria for content, concurrent, or criterion validity, as well as strategies to detect construct under- and overrepresentation, can provide valuable guidance for improving LLM benchmarks \citep{kalkbrenner2021practical,wongvorachan2025detecting}.

Alternatively, revising the nomological network amounts to revising the meaning of the construct itself. Researchers who rely on different nomological networks for reasoning implicitly operate with different conceptions of reasoning. Even well-established theories such as CHC may be challenged, particularly when applied to LLMs. As discussed above, CHC is ultimately a theory of \emph{human}, not \emph{LLM}, cognition and may require refinement.

Ultimately, nomological networks translate the meaning of capabilities into a set of statistical constraints on their benchmarks. If these constraints are satisfied, we can conclude that the empirical data cohere with the theory. One might object, following Borsboom, that “correlations are not enough, no matter what their size.” \citep{borsboom2004concept} Indeed, coherence alone does not establish Borsboom's high causal standards. But, if causal structures exist, systematic correlations will be among their observable consequences, rendering empirical coherence with the nomological network at least a necessary condition for its adequacy \citep{peters2017elements}. If no such causal structure exists, coherence between theory and data still provides partial justification for employing the theoretical vocabulary of human psychology when describing LLM capabilities.

\section{Conclusion}
\label{sec:conclusion}
The increasingly common practice of attributing human-like capabilities to LLMs on the basis of benchmark performance demands a rigorous theoretical foundation. I have argued that the nomological network account of construct validity from psychometrics offers the most promising path forward: it avoids contentious metaphysical commitments about the real existence of constructs in LLMs while providing a substantive and structured account of what those constructs mean. By embedding benchmarks within a theoretically articulated network of interrelated capabilities, we move from isolated performance scores to empirically constrained claims about LLM capabilities, as illustrated through the example of reasoning. If LLM research is to continue ascribing capabilities such as reasoning and theory of mind, it must also adopt the corresponding standards of validity that render these terms scientifically defensible.




\section*{Acknowledgements}\label{sec:acknow}
I thank Max Hellrigel-Holderbaum for his helpful comments on earlier drafts of this paper.

\bibliography{sn-bibliography}

\end{document}